\pdfoutput=1

\documentclass[11pt]{article}

\usepackage{ACL2023}

\usepackage{times}
\usepackage{latexsym}

\usepackage[T1]{fontenc}

\usepackage[utf8]{inputenc}

\usepackage{microtype}

\usepackage{inconsolata}

\usepackage{amsmath,amsfonts}
\usepackage{booktabs,arydshln}
\usepackage{multirow}
\usepackage{ulem}
\usepackage{pifont}
\usepackage{relsize}
\usepackage{enumitem}
\usepackage{graphicx}
\usepackage{caption,subcaption}
\usepackage{tikz}
\usepackage{mathtools}
\usepackage{fdsymbol}

\makeatletter
\def\adl@drawiv#1#2#3{%
        \hskip.5\tabcolsep
        \xleaders#3{#2.5\@tempdimb #1{1}#2.5\@tempdimb}%
                #2\z@ plus1fil minus1fil\relax
        \hskip.5\tabcolsep}
\newcommand{\cdashlinelr}[1]{%
  \noalign{\vskip\aboverulesep
           \global\let\@dashdrawstore\adl@draw
           \global\let\adl@draw\adl@drawiv}
  \cdashline{#1}
  \noalign{\global\let\adl@draw\@dashdrawstore
           \vskip\belowrulesep}}
\makeatother

\definecolor{approach}{HTML}{60d937}

\newcommand{\cmmnt}[1]{\ignorespaces}

\newcommand{\rem}[1]{\textrm{#1}}

\newcommand{\symbolsecref}[1]{($\S$~\ref{#1})}

\newcommand{\alignedmlm}[0]{\textsc{Align-mlm}}

\newcommand{\alignit}[0]{\textit{alignment}}
\newcommand{\alignitcapital}[0]{\textit{Alignment}}

\newcommand{\lang}[1]{$\mathcal{L}_{#1}$}
\newcommand{\orig}[0]{\lang{\textrm{NL}}}
\newcommand{\deriv}[0]{\lang{\textrm{deriv}}}

\newcommand{\corpus}[1]{\mathcal{C}_{#1}}

\newcommand{\mlm}[0]{\textsc{mlm}}
\newcommand{\tlm}[0]{\textsc{tlm}}
\newcommand{\xlm}[0]{\textsc{xlm}}
\newcommand{\dict}[0]{\textsc{dict-mlm}}

\newcommand{\trans}[0]{$\mathbf{\mathcal{T}}$}
\newcommand{\inv}[0]{$\mathbf{\mathcal{T}_{inv}}$}

\newcommand{\syntax}[0]{$\mathbf{\mathcal{T}_{syn}}$}
\newcommand{\script}[0]{$\mathbf{\mathcal{T}_{trans}}$}
\newcommand{\comp}[0]{$\mathbf{\circ}$}

\newcommand{\bz}[0]{\textit{ZS}}
\newcommand{\bs}[0]{\textit{SUP}}
\newcommand{\dlta}[0]{$\Delta$}

\title{\alignedmlm{}: Word Embedding Alignment is Crucial for Multilingual Pre-training}

\author{
Henry Tang \qquad Ameet Deshpande \qquad Karthik Narasimhan \\
Department of Computer Science\\
Princeton University, USA\\
\texttt{\{hrtang, asd, karthikn\}@cs.princeton.edu}
}

\begin{document}
\maketitle

\begin{abstract}

    
    Multilingual pre-trained models exhibit zero-shot cross-lingual transfer, where a model fine-tuned on a \textit{source} language achieves surprisingly good performance on a \textit{target} language.
    While studies have attempted to understand transfer, they focus only on MLM, and the large number of differences between natural languages makes it hard to disentangle the importance of different properties.
    In this work, we specifically highlight the importance of word embedding alignment by proposing a pre-training objective (\alignedmlm{}) whose auxiliary loss guides similar words in different languages to have similar word embeddings.
    \alignedmlm{} either outperforms or matches three widely adopted objectives (\mlm{}, \xlm{}, \dict{}) when we evaluate transfer between pairs of natural languages and their counterparts created by systematically modifying specific properties like the script.
    In particular,~\alignedmlm{} outperforms \xlm{} and \mlm{} by $35$ and $30$ F1 points on POS-tagging for transfer between languages that differ both in their script and word order (left-to-right v.s. right-to-left).
    We also show a strong correlation between alignment and transfer for all objectives (e.g., $\rho_{s}=0.727$ for XNLI), which together with~\alignedmlm{}'s strong performance calls for explicitly aligning word embeddings for multilingual models.
    \footnote{Code available at:\\ \url{https://github.com/princeton-nlp/align-mlm}}

\end{abstract}

\section{Introduction}
\label{sec:intro}


\begin{figure}[t]
    \centering
    \includegraphics[width=\linewidth]{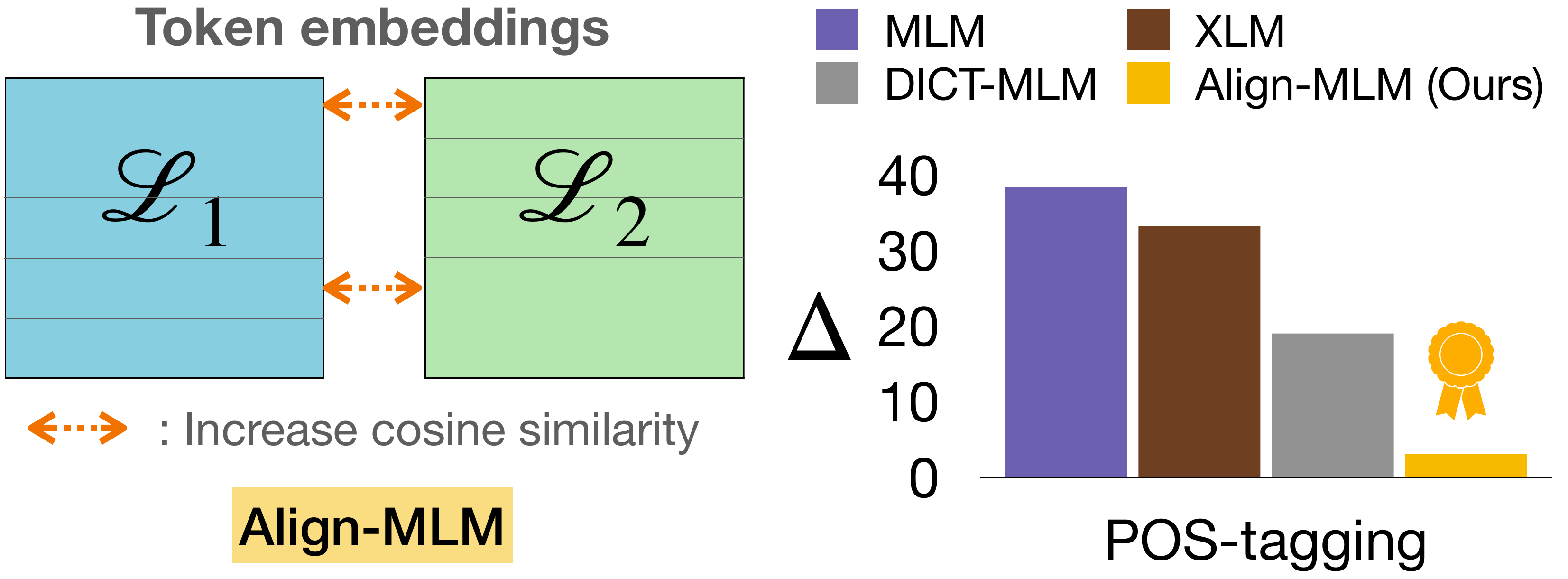}
    \caption{
    Our method \alignedmlm{} incorporates word embedding alignment by increasing the cosine similarity of embeddings corresponding to words in different languages with the same meaning (left).
    These words are chosen from a small bilingual dictionary (here, two word pairs denoted by orange arrows).
    Effectiveness of transfer is measured by the difference between supervised performance and zero-shot transfer (\dlta{}), with lower values being better.
    \alignedmlm{} (yellow) outperforms or matches other multilingual pre-training objectives (here by $>30$ F1 points on POS).
    }
    \label{fig:teaser}
\end{figure}


Multilingual pre-trained models like Multilingual-BERT~\cite{multilingualbert} and XLM~\cite{xlm-cite}
have shown impressive zero-shot cross-lingual transfer (hereon, transfer).
However, predicting the factors required for good transfer between languages is challenging, given the large number of linguistic factors which can vary between them (e.g., script, syntax, and vocabulary size).
Recent works have alleviated this issue~\cite{deshpande2021bert,ketal} by considering transfer between natural languages and counterparts constructed by modifying specific aspects (e.g., word order, transliteration).
However, they analyze only the masked language-modeling (MLM) pre-training objective, even though newer objectives like \xlm{}~\cite{xlm-cite} have been proposed and widely adopted.
Furthermore, while they highlight weaknesses of MLM, they do not propose improvements.

In this paper, we address both these issues by proposing a multilingual pre-training objective called \alignedmlm{} which highlights the importance of word embedding alignment.
\alignedmlm{} incorporates alignment across languages by increasing the cosine similarity of embeddings corresponding to words with the same meaning as part of an auxiliary loss.
\alignedmlm{} only requires access to a small bilingual dictionary, and does not require parallel sentences which are used by some objectives.
Analysis on three tasks (NLI, NER, POS) and three natural-derived language pairs from~\citet{deshpande2021bert} reveals that \alignedmlm{} always outperforms or performs on par with three widely used objectives: \mlm{}~\cite{devlin2019bert},  \xlm{} (uses parallel sentences)~\cite{xlm-cite} and \dict{} (uses a dictionary)~\cite{dict-mlm}.
For example, \alignedmlm{} outperforms \xlm{} and \mlm{} by $30$ and $35$ F1 points on part-of-speech tagging for a language pair which differ in their script and have inverted word orders.
\alignedmlm{}'s strong performance displays the importance of word embedding alignment for multilingual pre-training.

Furthermore, we show that even for other pre-training objectives (\mlm{}, \xlm{}, \dict{}), there exists a strong and positive correlation between word embedding alignment and zero-shot transfer ($\rho_{s}=0.727$ for XNLI).
This provides evidence that word embedding alignment is crucial for transfer, and its absence leads to poor transfer even for widely adopted objectives like \xlm{}.
\alignedmlm{} offers evidence that existing pre-training objectives can be improved by incorporating word alignment.

\section{Related Work}
\label{sec:related}


\paragraph{Analysis of multilingual models}
Multilingual pre-trained models~\cite{multilingualbert, xlm-cite,xlm-r,xue2021mt5,khanuja2021muril}
exhibit zero-shot cross-lingual transfer, but crucial properties required for transfer between pairs of languages is still unclear.
Several analysis studies have put forth inconsistent conclusions about factors like subword overlap and typological similarity~\cite{pires2019multilingual,conneau2020emerging,wu2019beto,hsu2019zero,lin2019choosing}.
Some recent studies~\cite{deshpande2021bert,wu2022oolong,dufter2020identifying,ketal} consider transfer in controlled settings, between natural languages and derived counterparts created by modifying specific linguistic aspects like script and word order.
However, these methods only investigate the masked language-modeling (MLM) objective~\cite{devlin2019bert}, whereas we additionally analyze newer pre-training methods like XLM~\cite{xlm-cite} and DICT-MLM~\cite{dict-mlm}.

\paragraph{Embedding alignment}
The \alignedmlm{} objective is inspired by our own analysis in this work and insight from previous studies that have explored word embedding alignment~\cite{cao2019multilingual,wang2019cross,schuster2019cross,ruder2019survey,yang2021bilingual,dict-mlm,khemchandani2021exploiting,dou2021word}.
\alignedmlm{} differs from these studies by (1) using bilingual dictionaries instead parallel sentences, (2) applying our objective during pre-training, and (3) explicitly aligning the embedding spaces.



\begin{figure*}[ht]

    \centering
    \begin{subfigure}[b]{0.334\textwidth}
    \centering
    \includegraphics[width=\textwidth]{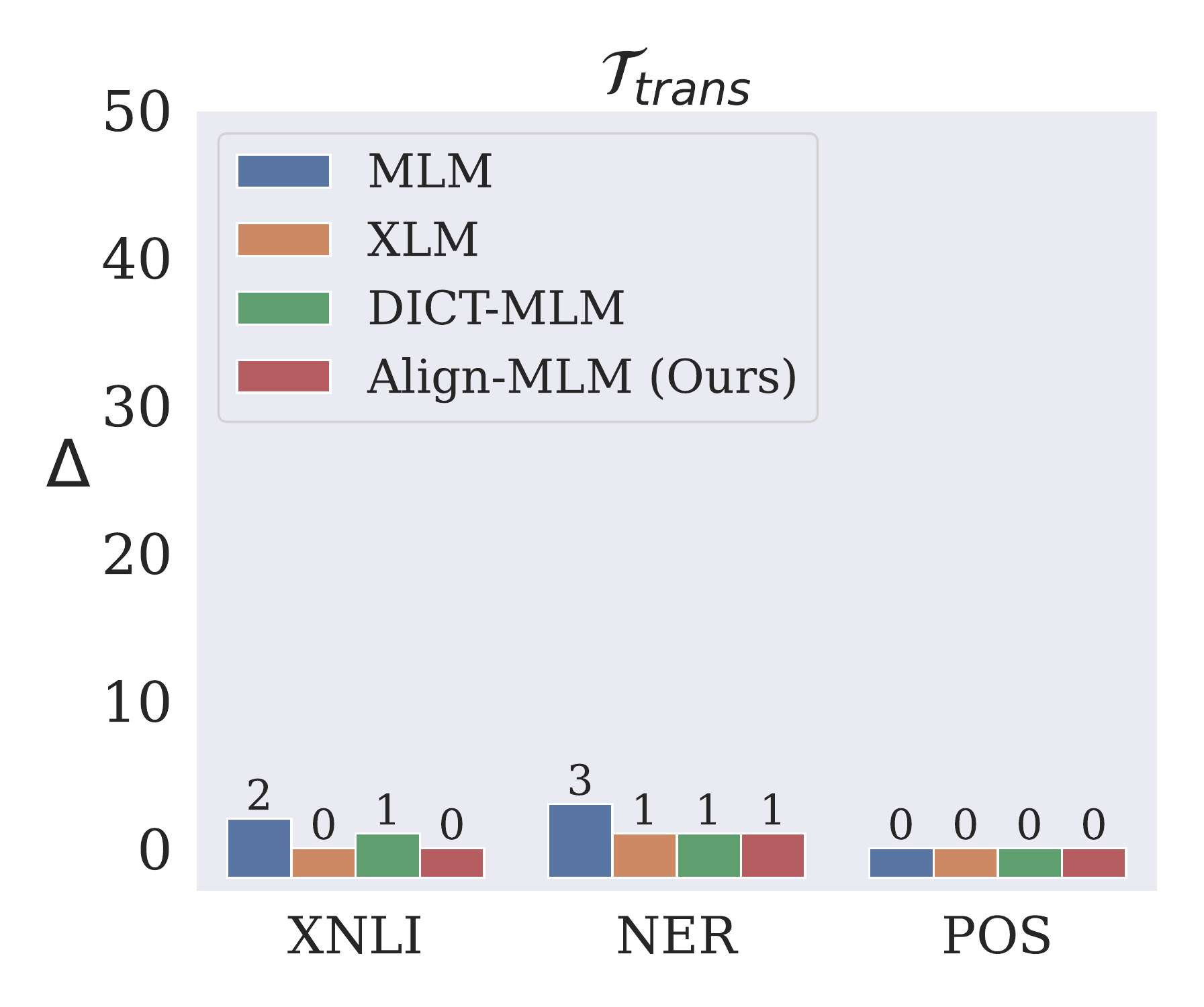}
    \end{subfigure}
    \begin{subfigure}[b]{0.326\textwidth}
    \centering
    \includegraphics[width=\textwidth]{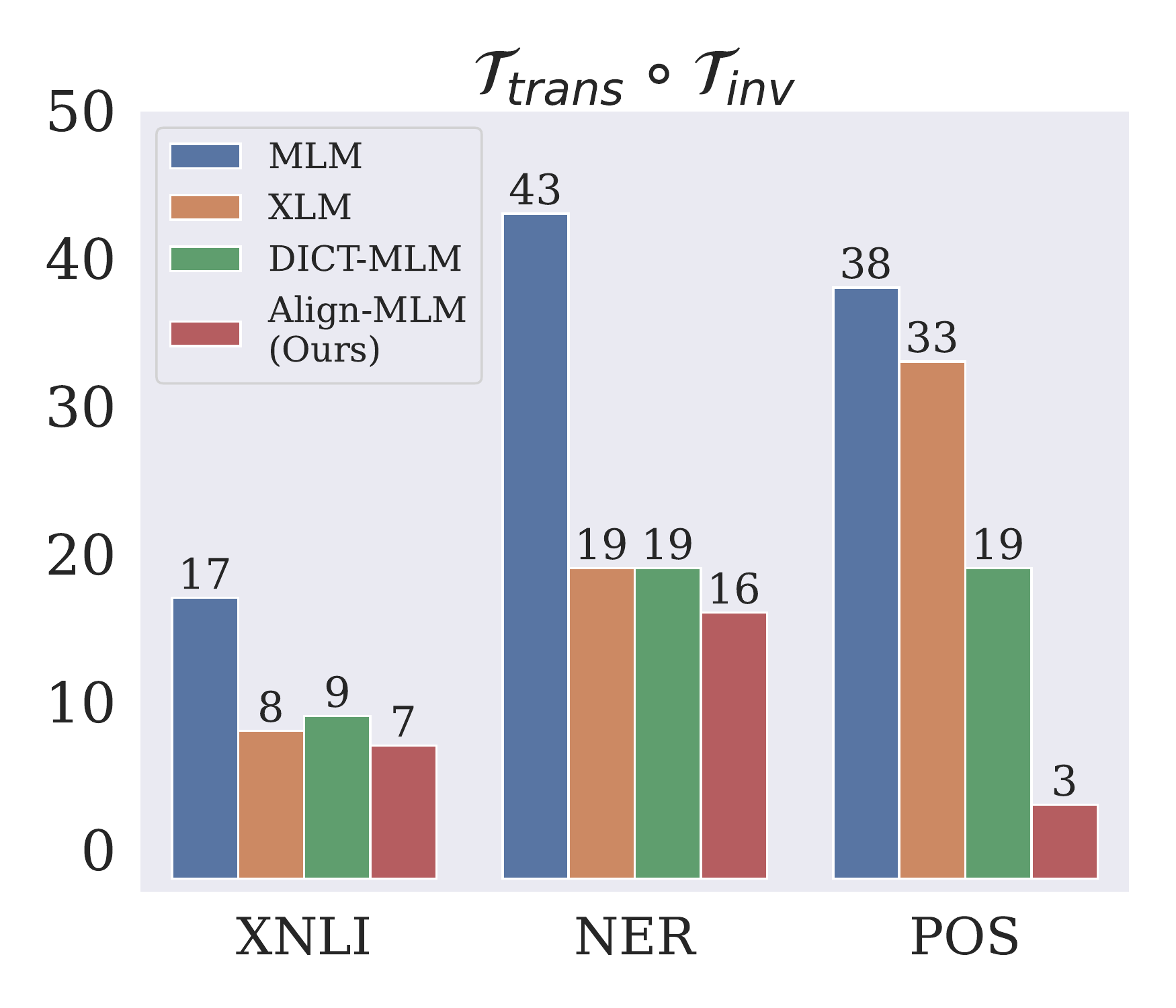}
    \end{subfigure}
    \begin{subfigure}[b]{0.326\textwidth}
    \centering
    \includegraphics[width=\textwidth]{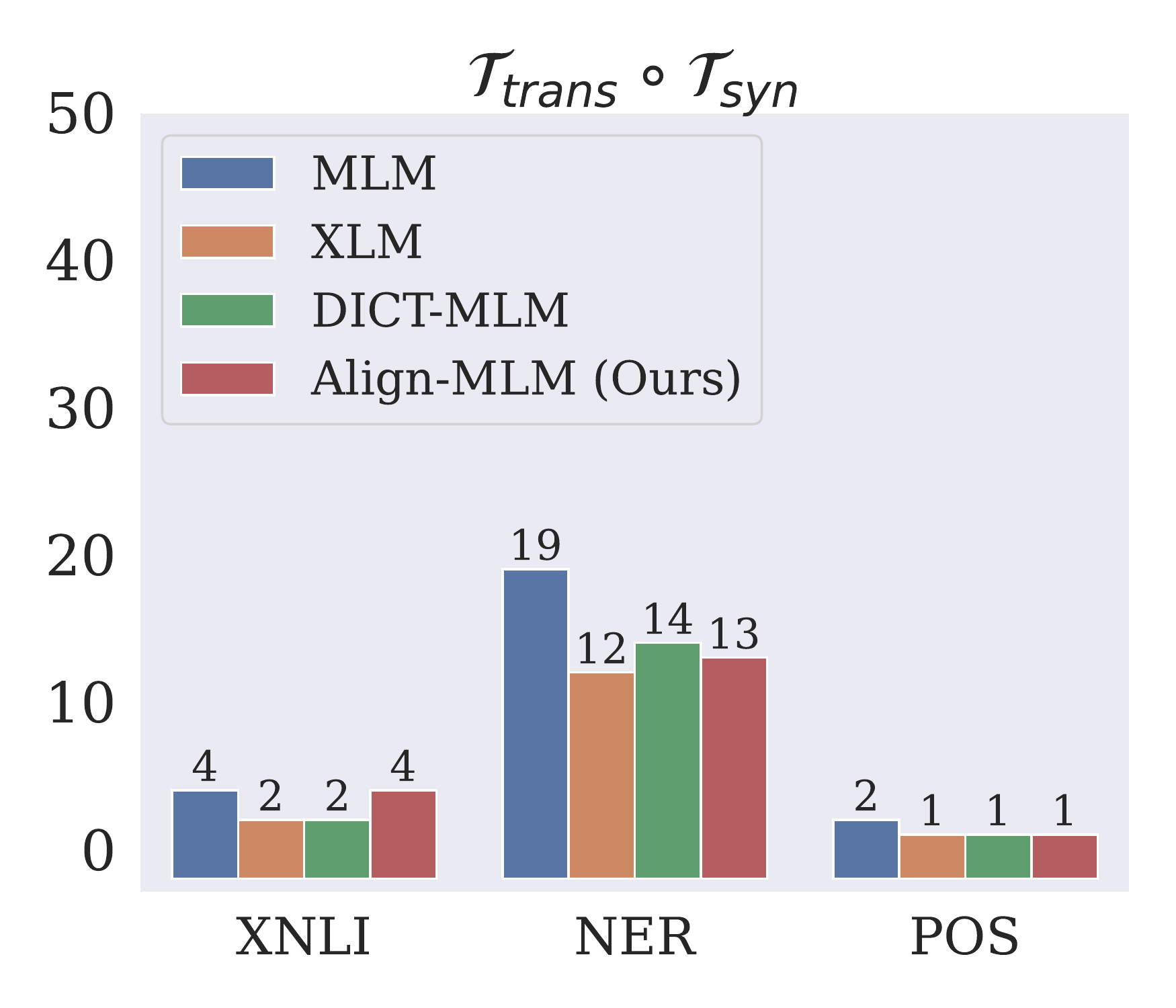}
    \end{subfigure}
    
    \caption{
    We compare \alignedmlm{} (red) with other pre-training objectives on three transformations~\symbolsecref{sec:results} and observe that it significantly outperforms them for \script{} \comp{} \inv{} (second plot).
    For example,~\alignedmlm{} is better than XLM by $30$ points on POS and MLM by $10$ points on XNLI.
    \alignedmlm{} also matches the performance of XLM on \script{} \comp{} \syntax{} without using any parallel sentences, and outperforms MLM by over $6$ points on NER.
    }
    \label{fig:main_results}
    
\end{figure*}


    
    

\section{Approach}
\label{sec:approach}

We analyze multilingual models pre-trained on two languages (\lang{1}, \lang{2}) and describe the components.

\paragraph{Multilingual pre-training}
We consider three existing multilingual pre-training objectives.
(1) Masked language-modeling (\mlm{})~\cite{devlin2019bert} involves pre-training a Transformer~\cite{vaswani2017attention} simultaneously on both languages.
(2) Cross-lingual language models (\xlm{})~\cite{xlm-cite} additionally use translation language-modeling (\tlm{}) which performs \mlm{} on a corpus of parallel sentence pairs.
Formally $\corpus{\rem{1}}$, $\corpus{\rem{2}}$ are the corpora for \lang{1} and \lang{2} respectively, and $\corpus{\rem{1--2}}$ contains the parallel sentences.
(3) \dict{}~\cite{dict-mlm} uses \mlm{}, but it randomly translates tokens in masked positions to the other language with the help of a bilingual dictionary, thus creating code-switched data ($\tilde{\mathcal{C}}_{\rem{1}}$)~\cite{newman1951language}.
\vspace{-0.5em}
\begin{align*}
    &\textrm{L}_{\textrm{\textbf{\mlm{}}}} \coloneqq \textrm{L}_{\rem{\mlm{}}}\left ( \corpus{\rem{1}}  \right ) + \textrm{L}_{\rem{\mlm{}}}\left ( \corpus{\rem{2}}  \right ) \\
    &\textrm{L}_{\textrm{\textbf{\xlm{}}}} \coloneqq \textrm{L}_{\rem{\mlm{}}}\left ( \corpus{\rem{1}}  \right ) + \textrm{L}_{\rem{\mlm{}}}\left ( \corpus{\rem{2}}  \right ) + \textrm{L}_{\rem{\tlm{}}}\left ( \corpus{\rem{1--2}}  \right ) \\    
    &\textrm{L}_{\textrm{\textbf{\dict{}}}} \coloneqq \textrm{L}_{\rem{\mlm{}}}\left ( \tilde{\mathcal{C}}_{\rem{1}}  \right ) + \textrm{L}_{\rem{\mlm{}}}\left ( \tilde{\mathcal{C}}_{\rem{2}}  \right )
\end{align*}

\paragraph{Incorporating alignment during pre-training}
We propose \alignedmlm{} which incorporates word embedding alignment by explicitly encouraging it during pre-training through an auxiliary loss, and compare it with the aforementioned objectives.
\alignedmlm{} uses a bilingual dictionary to guide word embeddings of translations to be similar by increasing their cosine similarity.
Formally, let $\mathcal{B}$ be a bilingual dictionary with $\mathcal{B}_{i1}$ and $\mathcal{B}_{i2}$, belonging to \lang{1} and \lang{2}, corresponding to the tokens in the $\textrm{i}^{\textrm{th}}$ entry.
Further, let $\mathcal{E}[\mathcal{B}_{ij}]$ be the embedding of token $\mathcal{B}_{ij}$.
$\alpha$ is a hyperparameter which controls the relative importance of the alignment loss.
\begin{align*}
    &\textrm{L}_{\textrm{\textsc{align}}} \coloneqq -\dfrac{1}{|\mathcal{B}|} \sum_{i=1}^{|\mathcal{B}|} \textrm{\textit{cos sim}}\left ( \mathcal{E}[\mathcal{B}_{i1}], \mathcal{E}[\mathcal{B}_{i2}]  \right )\\
    &\textrm{L}_{\textrm{\textbf{\alignedmlm{}}}} \coloneqq \textrm{L}_{\textrm{\textbf{\mlm{}}}} + \alpha \cdot \textrm{L}_{\textrm{\textsc{align}}}
\end{align*}
Unlike \xlm{}, \alignedmlm{} does not use parallel sentences, but only a bilingual dictionary which is easy to obtain and also used by \dict{}.



\paragraph{Analysis setup}

We compare pre-training objectives on systematically generated language pairs which have realistic and controlled differences generated by~\citet{deshpande2021bert}.
We use three transformations to modify properties of a \textit{natural} language (\orig{}) and create a \textit{derived} language \deriv{}: (1) transliteration -- \script{} (changes the script of the language), (2) inversion -- \inv{} (inverts the order of tokens in the sentence), and (3) syntax -- \syntax{} (modifies the syntactic properties, namely the subject-verb-object and noun-adjective order).
The transformations are applied both at pre-training and fine-tuning time to create corresponding \textit{derived} language datasets, and we present the procedure and examples in Appendix~\ref{app:transformations}.
We pre-train a separate model for every natural-derived language pair.
For each pair, we measure zero-shot transfer (\bz{})
by fine-tuning the model on a downstream task in \orig{} and testing on \deriv{}.
We also compute the supervised performance (\bs{}) by both fine-tuning and testing on \deriv{}.
Since \bs{} serves as a plausible upper-bound on \bz{}, a smaller difference (\dlta = $\textrm{\bs{}} - \textrm{\bz{}}$) characterizes better transfer.

\section{Experimental setup}
We consider three language pairs with English as \orig{}.
\deriv{} is created by applying the transformations (1) \script{}, (2) \script{} \comp{} \inv{}, and (3) \script{} \comp{} \syntax{}, where $\mathbf{\mathcal{T}_{t_1}}$ \comp{} $\mathbf{\mathcal{T}_{t_2}}$ represents applying transformation $\mathbf{\mathcal{T}_{t_2}}$ followed by $\mathbf{\mathcal{T}_{t_1}}$.
Thus, \script{} \comp{} \syntax{} creates a language which differs from English both in its script and syntax.
We evaluate on three tasks from XTREME~\cite{hu2020xtreme} -- natural language inference~\cite{conneau2018xnli}, named-entity recognition~\cite{pan2017cross}, and part-of-speech tagging~\cite{ud}, with accuracy for NLI and F1 for NER and POS as the metrics.
We compare \mlm{}, \xlm{}, \dict{}, and \alignedmlm{} on all language pairs and tasks, and defer pre-training and other details to Appendices~\ref{app:implementation},~\ref{app:hyperparameters}.

\section{Results}
\label{sec:results}

\paragraph{ALIGN-MLM outperforms MLM, XLM, and DICT-MLM}



\alignedmlm{}'s strong performance highlights the importance of word embedding alignment (Figure~\ref{fig:main_results}).
All objectives transfer well between language pairs which differ only in their script (\script{}), as substantiated by \dlta{} $\approx 0$ (first plot, Fig~\ref{fig:main_results}).
However, \mlm{}, \xlm{}, and \dict{} perform poorly on language pairs which differ in their script and have inverted word orders (\script{} \comp{} \inv{}, second plot--Fig~\ref{fig:main_results}), with \alignedmlm{} outperforming or matching them.
For example, \alignedmlm{} is better than both \xlm{} and \dict{} by over $1.5$ and $3.5$ points on XNLI and NER.
The difference is even more significant for POS tagging, with \alignedmlm{} outperforming \xlm{} by over $30$ points.
\alignedmlm{}'s zero-shot score (\bz{}) is $92.0$, as opposed to \xlm{}'s $61.8$ and \dict{}'s $76.0$.
For \script{} \comp{} \syntax{}, which makes local syntax changes when compared to \script{} \comp{} \inv{}, we observe that all objectives outperform \mlm{} (third plot, Fig~\ref{fig:main_results}) with \alignedmlm{} beating \mlm{} by over $6$ points on NER.
\alignedmlm{}'s strong performance when compared to prior widely adopted objectives underlines the importance of alignment.



\paragraph{Does more parallel data help \xlm{}?}

\begin{table}[t]
    \centering
    \resizebox{\columnwidth}{!}{
    \begin{tabular}{crrrrrr}
    \toprule
    \bf{\# parallel sent} & \multicolumn{2}{c}{\bf{XNLI}} & \multicolumn{2}{c}{\bf{NER}} & \multicolumn{2}{c}{\bf{POS}} \\
    \cmidrule(lr){2-3} \cmidrule(lr){4-5} \cmidrule(lr){6-7}
    \bf{rel. to \mlm{}} & \bz{} $\uparrow$ & \dlta{} $\downarrow$ & \bz{} $\uparrow$ & \dlta{} $\downarrow$ & \bz{} $\uparrow$ & \dlta{} $\downarrow$ \\ 
    \midrule
    1\% & 59.8                   & 16.1                      & 40.2                   & 39.5                      & 66.9                   & 28.1                      \\
    \midrule
    10\% & 65.8                   & 9.5                       & 59.4                   & 20.3                    & 63.6                   & 31.5                      \\ 
    \midrule
    25\% & 67.2                   & 8.2                       & 60.7                   & 19.4                    & 61.8                   & 33.2                      \\
    \midrule
    100\% & 65.2                   & 9.8                       & 62.4                   & 17.6                    & 60.0                   & 35.1                      \\ \cdashlinelr{1-7}
    \alignedmlm{} (0\%) & \bf{68.8} & \bf{6.7} & \bf{64.5} & \bf{15.5} & \bf{92.0} & \bf{3.0} \\
    \bottomrule
    \end{tabular}
    }
    \caption{For \script{} \comp{} \inv{}, we vary the number of parallel sentences used by \xlm{} relative to the size of the unsupervised corpus used by \mlm{} (\%).
    \alignedmlm{}, which does not use parallel sentences and only a dictionary, outperforms \xlm{} on all tasks even when it uses a large parallel corpus (100\%).}
    \label{tab:xlm_analysis}
\end{table}

Results in Figure~\ref{fig:main_results} show that \alignedmlm{} outperforms \xlm{} even though the former does not use parallel data.
We now vary the amount of parallel sentences used by \xlm{} relative to the number of sentences used for standard \mlm{} pre-training.
For example, if the percentage of parallel sentences is 25\% and we use 10 million sentences for \mlm{} pre-training, we use 2.5M parallel sentences in \xlm{}.
We note from our results in Table~\ref{tab:xlm_analysis}, that~\alignedmlm{} outperforms all versions of \xlm{}, even though it does not use any parallel sentences and uses only a bilingual dictionary with 25\% of the tokens in the vocabulary.
In fact, even in the unrealistic case of equal amounts of parallel and \mlm{} data (100\%), \xlm{} performs worse than \alignedmlm{} by $\approx 3, 2, 32$ points on XNLI, NER, and POS respectively.
This shows that simply increasing the amount of parallel data does not improve alignment, and explicit objectives like~\alignedmlm{} perform better.

\paragraph{\dict{} v.s. \alignedmlm{}}

\newcommand{\alignedmlmnew}[0]{{\footnotesize \textsc{Align}}}
\newcommand{\dictnew}[0]{{\footnotesize \textsc{dict}}}

\begin{table}[t]
    \centering
    \resizebox{\columnwidth}{!}{
    \begin{tabular}{crrrrrr}
    \toprule
    \bf{\% tokens in} & \multicolumn{2}{c}{\bf{XNLI~(\dlta{} $\downarrow$)}} & \multicolumn{2}{c}{\bf{NER~(\dlta{} $\downarrow$)}} & \multicolumn{2}{c}{\bf{POS~(\dlta{} $\downarrow$)}} \\
    \cmidrule(lr){2-3} \cmidrule(lr){4-5} \cmidrule(lr){6-7}
    \bf{dictionary} & \dictnew{} & \alignedmlmnew{} & \dictnew{} & \alignedmlmnew{} & \dictnew{} & \alignedmlmnew{} \\
    \midrule
    25\% & 9.2 & \bf{6.7} & 19.3 & \bf{15.5} & 19.0 & \bf{3.0} \\
    50\% & 6.0 & \bf{5.7} & 15.9 & \bf{14.0} & 3.4 & \bf{1.5} \\
    \bottomrule
    \end{tabular}
    }
    \caption{For \script{} \comp{} \inv{}, we vary the number of tokens in the bilingual dictionary relative the the vocabulary size of \orig{}.
    \alignedmlm{} outperforms \dict{} on all tasks for both settings (25\% and 50\%), implying that~\alignedmlm{} uses dictionaries more efficiently.
    }
    \label{tab:dict_vs_aligned}
\end{table}

We vary the size of the bilingual dicionary used by these methods as a percentage relative to the vocabulary size of \orig{} and compare their performance in Table~\ref{tab:dict_vs_aligned} (due to space constraints we report results for \script{} \comp{} \syntax{} in Appendix~\ref{app:dict_vs_aligned}).
We note that for all the tasks and for both 25\% and 50\% of the tokens in the dictionary, \alignedmlm{} outperforms \dict{}.
Moreover, this is true both for \script{} \comp{} \inv{} and \script{} \comp{} \syntax{}, which shows that \alignedmlm{}'s explicit alignment might be a better strategy than creating code-switched data like in~\dict{}.


\paragraph{Correlation between word embedding alignment and transfer}
\begin{figure}[t]
    \centering
    \includegraphics[width=\linewidth]{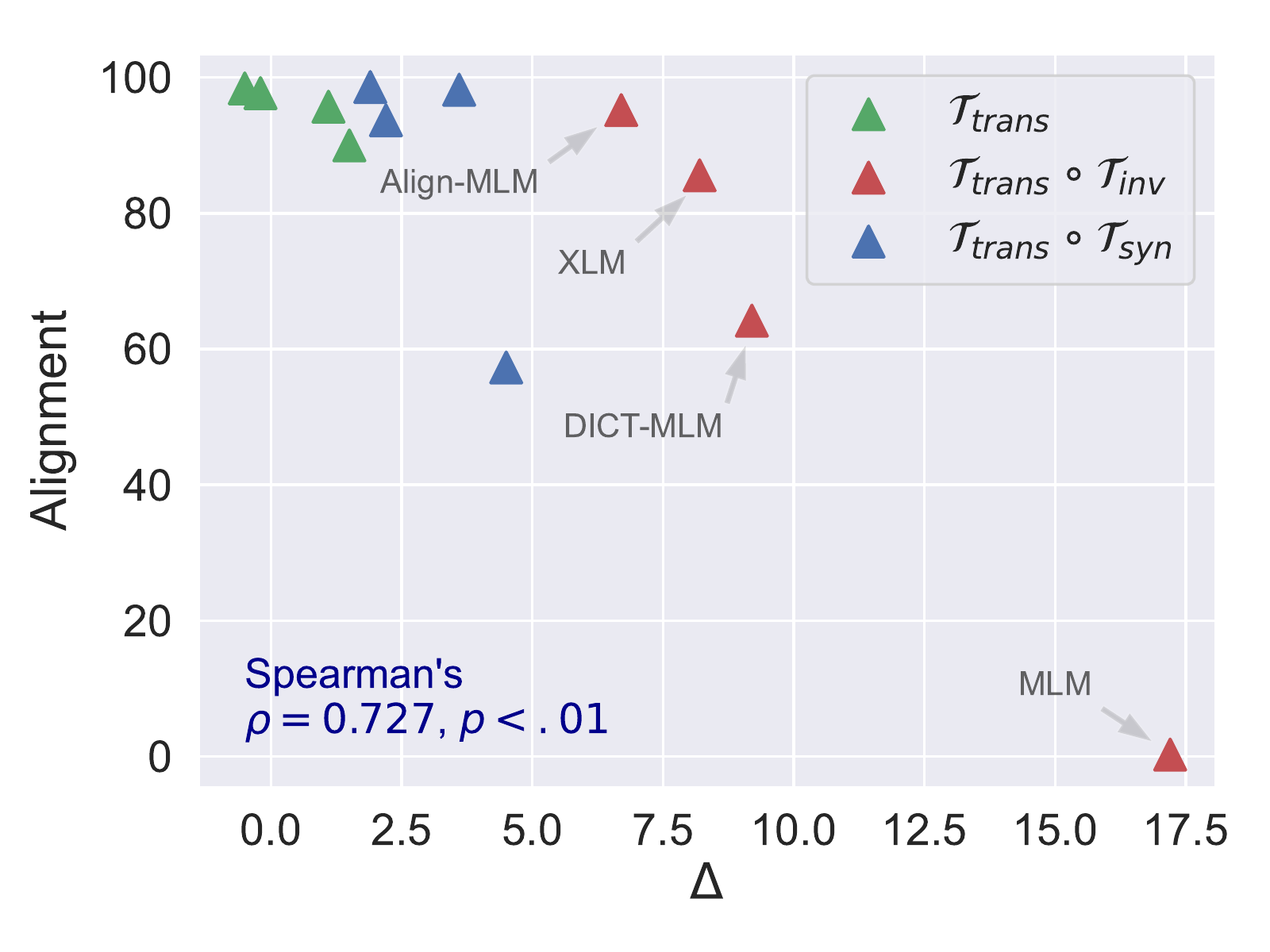}
    \caption{
    \alignitcapital{} vs \dlta{} for XNLI. We see that better transfer (lower \dlta{}) is strongly correlated with better \alignit{} ($\rho_{s}=0.73$).
    For example, for \script{} \comp{} \inv{}, \alignedmlm{}, has both the best transfer and \alignit{}.
    }
    \label{fig:xnli_correlation_annotated}
\end{figure}
There exists a one-to-one correspondence between the vocabularies of \orig{} and \deriv{} when transliteration (\script{}) is used (even if other transformations are used in addition).
Following~\citet{deshpande2021bert}, the \textit{alignment score} of a token in the vocabulary of \orig{} can be defined as 100\% if its cosine similarity with the corresponding token in the \deriv{} is highest among \deriv{}'s vocabulary, else it is 0\%.
We denote the average alignment for all tokens in the \orig{} to be ``\alignit{}''.
Figure~\ref{fig:xnli_correlation_annotated} shows that higher \alignit{} is correlated with better transfer (smaller \dlta{}) on XNLI.
For example, across pre-training methods, for language pairs constructed using \script{} \comp{} \inv{}, better \alignit{} implies better transfer (red triangles).
Better \alignit{} is strongly correlated with better transfer as measured by Spearman correlation ($\rho_{s} = 0.73$, $p\textrm{ {\footnotesize (2-tailed)}}<.01$), and this is also true for other tasks (Appendix~\ref{app:correlation}).
This highlights the importance of alignment for all pre-training methods.



\section{Conclusion}
\label{sec:conclusion}
\alignedmlm{} highlights the importance of word embedding alignment during pre-training.
It matches or even outperforms several objectives from prior work (\mlm{}, \xlm{}, \dict{}).
We also show that \alignit{} is strongly correlated with better transfer performance for all objectives.
Taken together, our results call for improvements in multilingual pre-training objectives, and we recommend doing so by explicitly incorporating alignment, rather than relying on its emergence.

\newpage
\section{Limitations}
Our focus for this paper is on creating a controlled study because previous studies have failed to conclude anything significant while using natural languages (refer to discussion in~\symbolsecref{sec:related}). Our experiments are done on language pairs which have controlled differences, because natural languages can have multiple differences which makes it hard to pinpoint crucial properties for transfer.
While we are comprehensive in creating these language pairs and use ones in literature~\cite{deshpande2021bert}, future studies can consider more properties while creating language pairs.
Further, we need large scale compute and use TPUs for our analysis because of the scale of our experiments.

\section*{Acknowledgement}
This work was funded through a grant from the Chadha Center for Global India at Princeton University.
We thank the Google Cloud Research program for computational support.


\bibliography{anthology,custom}
\bibliographystyle{acl_natbib}

\newpage
\appendix
\section*{Appendix}

\section{Comparing different pre-training objectives}
\label{app:comparing}

We add an extended version of Figure~\ref{fig:main_results} which compares different pre-training objectives in Table~\ref{app:tab:main_table}.

\begin{table*}[ht]
    \centering
    \resizebox{2\columnwidth}{!}{
    \begin{tabular}{clrrrrrrrrr}
    \toprule
    \multicolumn{1}{l}{\bf{Pre-training method}} & \bf{Transformation} & \multicolumn{3}{c}{\bf{XNLI}}                                                                                                    & \multicolumn{3}{c}{\bf{NER}}                                                                                                     & \multicolumn{3}{c}{\bf{POS}}                                                                                                     \\ \cmidrule(lr){3-5} \cmidrule(lr){6-8} \cmidrule(lr){9-11}
    \multicolumn{1}{l}{}                    &                & \multicolumn{1}{l}{\bs{}} & \multicolumn{1}{l}{\bz{}} & \multicolumn{1}{l}{\dlta{}} & \multicolumn{1}{l}{\bs{}} & \multicolumn{1}{l}{\bz{}} & \multicolumn{1}{l}{\dlta{}} & \multicolumn{1}{l}{\bs{}} & \multicolumn{1}{l}{\bz{}} & \multicolumn{1}{l}{\dlta{}} \\ \midrule
    & Trans          & 76.4                     & 74.9                   & 1.5                       & 80.6                    & 78.0                   & 2.6                       & 95.0                    & 94.6                   & 0.4                       \\
                        & Trans + Inv    & 73.9                     & 56.7                   & 17.2                      & 79.2                    & 35.9                   & 43.3                      & 95.0                    & 56.5                   & 38.5                      \\
    \multirow{-3}{*}{\mlm{}} & Trans + Syn    & 73.5                     & 69.0                   & 4.5                       & 71.3                    & 52.3                   & 19.0                      & 93.6                    & 91.6                   & 2.0                       \\
    \midrule
    & Trans          & 77.3                     & 77.5                   & -0.2                      & 80.8                    & 79.7                   & 1.1                       & 95.0                    & 94.8                   & 0.2                       \\
                        & Trans + Inv    & 75.4                     & 67.2                   & 8.2                       & 80.0                    & 60.7                   & 19.4                      & 95.0                    & 61.8                   & 33.2                      \\
    \multirow{-3}{*}{\xlm{}}                    & Trans + Syn    & 76.1                     & 74.1                   & 1.9                       & 71.5                    & 59.6                   & 11.9                      & 93.7                    & 92.7                   & 0.9                       \\
    \midrule
    & Trans          & 77.0                     & 75.9                   & 1.1                       & 79.3                    & 78.3                   & 1.1                       & 95.1                    & 94.8                   & 0.3                       \\
                        & Trans + Inv    & 74.5                     & 65.3                   & 9.2                       & 79.1                    & 59.8                   & 19.3                      & 95.0                    & 76.0                   & 19.0                      \\
    \multirow{-3}{*}{\dict{}}                    & Trans + Syn    & 74.3                     & 72.2                   & 2.2                       & 71.4                    & 57.6                   & 13.8                      & 93.6                    & 92.6                   & 1.0                       \\
    \midrule
    & Trans          & 75.1                     & 75.5                   & -0.5                      & 79.4                    & 78.6                   & 0.8                       & 94.9                    & 94.8                   & 0.1                       \\
                        & Trans + Inv    & 75.5                     & 68.8                   & 6.7                       & 80.0                    & 64.5                   & 15.5                      & 95.0                    & 92.0                   & 3.0                       \\
    \multirow{-3}{*}{\alignedmlm{}}                    & Trans + Syn    & 74.7                     & 71.1                   & 3.6                       & 71.3                    & 58.3                   & 13.1                      & 93.7                    & 92.9                   & 0.8                       \\ \bottomrule
    \end{tabular}
    }
    \caption{Results for all four pre-training methods considered in the paper. All four pre-training methods use their default setup.}
    \label{app:tab:main_table}
    \end{table*}

\section{\dict{} vs \alignedmlm{}}
\label{app:dict_vs_aligned}

We compare \dict{} and \alignedmlm{} while varying the size of the bilingual dictionary relative to the vocabulary size of \orig{} and report the results in tables~\ref{app:tab:dict_vs_aligned} and~\ref{app:tab:dict_vs_aligned_syntax} for \script{} \comp{} \inv{} and \script{} \comp{} \syntax{} respectively.
We observe that in all but one case~\alignedmlm{} outperforms \dict{} by making better utilization of the bilingual dictionary.
\newcommand{\alignedmlmnewapp}[0]{{\footnotesize \textsc{Align}}}
\newcommand{\dictnewapp}[0]{{\footnotesize \textsc{dict}}}

\begin{table}[ht]
    \centering
    \resizebox{\columnwidth}{!}{
    \begin{tabular}{crrrrrr}
    \toprule
    \bf{\% tokens in} & \multicolumn{2}{c}{\bf{XNLI~(\dlta{})}} & \multicolumn{2}{c}{\bf{NER~(\dlta{})}} & \multicolumn{2}{c}{\bf{POS~(\dlta{})}} \\
    \cmidrule(lr){2-3} \cmidrule(lr){4-5} \cmidrule(lr){6-7}
    \bf{dictionary} & \dictnewapp{} & \alignedmlmnewapp{} & \dictnewapp{} & \alignedmlmnewapp{} & \dictnewapp{} & \alignedmlmnewapp{} \\
    \midrule
    25\% & 9.2 & \bf{6.7} & 19.3 & \bf{15.5} & 19.0 & \bf{3.0} \\
    50\% & 6.0 & \bf{5.7} & 15.9 & \bf{14.0} & 3.4 & \bf{1.5} \\
    \bottomrule
    \end{tabular}
    }
    \caption{Results for \script{} \comp{} \inv{}}
    \label{app:tab:dict_vs_aligned}
\end{table}

\begin{table}[ht]
    \centering
    \resizebox{\columnwidth}{!}{
    \begin{tabular}{crrrrrr}
    \toprule
    \bf{\% tokens in} & \multicolumn{2}{c}{\bf{XNLI~(\dlta{})}} & \multicolumn{2}{c}{\bf{NER~(\dlta{})}} & \multicolumn{2}{c}{\bf{POS~(\dlta{})}} \\
    \cmidrule(lr){2-3} \cmidrule(lr){4-5} \cmidrule(lr){6-7}
    \bf{dictionary} & \dictnewapp{} & \alignedmlmnewapp{} & \dictnewapp{} & \alignedmlmnewapp{} & \dictnewapp{} & \alignedmlmnewapp{} \\
    \midrule
    25\% & \bf{2.2} & 3.6 & 13.8 & \bf{13.1} & 1.0 & \bf{0.8} \\
    50\% & 2.9 & \bf{1.6} & 13.4 & \bf{11.5} & 0.9 & \bf{0.7} \\
    \bottomrule
    \end{tabular}
    }
    \caption{Results for \script{} \comp{} \syntax{}}
    \label{app:tab:dict_vs_aligned_syntax}
\end{table}

\section{Varying the amount of parallel data for \xlm{}}
\label{app:tlm_data}

We vary the amount of parallel data used by \xlm{} relative to the amound of unsupervised data used by \mlm{} and report the results in Table~\ref{app:tab:tlm}.
We notice that as we increase the amount of parallel sentences, performance improves for XNLI and NER, but does not change or decreases for POS.
Regardless of the task, transfer can be poor even when a large number of parallel sentences are used (such as 100\%), as measured by \dlta{}.

\begin{table*}[ht]
\centering
\begin{tabular}{clrrrrrrrrr}
\toprule
\multicolumn{1}{c}{\bf{\# parallel sent}} & \multicolumn{3}{c}{\bf{XNLI}} & \multicolumn{3}{c}{\bf{NER}} & \multicolumn{3}{c}{\bf{POS}} \\
\cmidrule(lr){2-4} \cmidrule(lr){5-7} \cmidrule(lr){8-10}
\multicolumn{1}{c}{\bf{relative to \mlm{}}} & \multicolumn{1}{l}{\bs{}} & \multicolumn{1}{l}{\bz{}} & \multicolumn{1}{l}{\dlta{}} & \multicolumn{1}{l}{\bs{}} & \multicolumn{1}{l}{\bz{}} & \multicolumn{1}{l}{\dlta{}} & \multicolumn{1}{l}{\bs{}} & \multicolumn{1}{l}{\bz{}} & \multicolumn{1}{l}{\dlta{}} \\ 
\midrule
1\% & 75.9                     & 59.8                   & 16.1                      & 79.8                    & 40.2                   & 39.5                      & 95.0                    & 66.9                   & 28.1                      \\
\midrule
10\% & 75.3                     & 65.8                   & 9.5                       & 79.7                    & 59.4                   & 20.3                      & 95.1                    & 63.6                   & 31.5                      \\ 
\midrule
25\% & 75.4                     & 67.2                   & 8.2                       & 80.0                    & 60.7                   & 19.4                      & 95.0                    & 61.8                   & 33.2                      \\
\midrule
100\% & 75.0                     & 65.2                   & 9.8                       & 80.0                    & 62.4                   & 17.6                      & 95.1                    & 60.0                   & 35.1                      \\
\bottomrule
\end{tabular}
\caption{\xlm{} results when varying the amount of parallel sentences for \script{} \comp{} \inv{}.}
\label{app:tab:tlm}
\end{table*}

\section{Vocabulary overlap}
\label{app:vocabulary}

We measure the performance of \xlm{} when \orig{} and \deriv{} share a percentage of the vocabulary for \script{} \comp{} \inv{}.
Instead of \script{} changing the script of all the tokens in \orig{}, it changes only a percentage of tokens.
We present the results in Table~\ref{app:tab:vocabulary}, where we notice that as \orig{} and \deriv{} share more sub-words, transfer improves, as expected.

\begin{table*}[ht]
\centering
\begin{tabular}{clrrrrrrrrr}
\toprule
\multicolumn{1}{c}{\bf{\% vocabulary}} & \multicolumn{3}{c}{\bf{XNLI}} & \multicolumn{3}{c}{\bf{NER}} & \multicolumn{3}{c}{\bf{POS}} \\
\cmidrule(lr){2-4} \cmidrule(lr){5-7} \cmidrule(lr){8-10}
\multicolumn{1}{c}{\bf{overlap}} & \multicolumn{1}{l}{\bs{}} & \multicolumn{1}{l}{\bz{}} & \multicolumn{1}{l}{\dlta{}} & \multicolumn{1}{l}{\bs{}} & \multicolumn{1}{l}{\bz{}} & \multicolumn{1}{l}{\dlta{}} & \multicolumn{1}{l}{\bs{}} & \multicolumn{1}{l}{\bz{}} & \multicolumn{1}{l}{\dlta{}}\\
\midrule
0\% & 75.4 & 67.2 & 8.2   & 80.0 & 60.7 & 19.4  & 95.0 & 61.8 & 33.2  \\
\midrule
25\% & 76.6 & 71.3 & 5.2   & 79.8 & 64.0 & 15.8  & 95.1 & 92.6 & 2.4   \\
\midrule
50\% & 75.0 & 72.2 & 2.8   & 80.0 & 42.5 & 37.5  & 95.0 & 94.4 & 0.6   \\
\bottomrule
\end{tabular}
\caption{\xlm{} results when varying the percentage of vocabulary overlap between \lang{1} and \lang{2}, for the transformation \script{} \comp{} \inv{}.}
\label{app:tab:vocabulary}
\end{table*}

\section{Analyzing word embedding alignment and zero-shot transfer}
\label{app:correlation}

\begin{table*}[ht]
\centering
\begin{tabular}{clrrrrrrrrr}
\toprule
\multicolumn{1}{l}{\bf{Pre-training method}}
& \multicolumn{1}{l}{\bf{Transformation}} & \multicolumn{1}{l}{\bf{Alignment}} & \multicolumn{1}{l}{\bf{XNLI \dlta{} ($\downarrow$)}} \\
\midrule
\multirow{3}{*}{\mlm{}}                    & Trans          & 90.0 & 1.5                        \\
                                        & Trans + Inv    & 0.3  & 17.2                       \\
                                        & Trans + Syn    & 57.3 & 4.5                        \\
\midrule
\multirow{3}{*}{\xlm{}}                    & Trans          & 97.7 & 0.2                         \\
                                        & Trans + Inv    & 85.6 & 8.2                         \\
                                        & Trans + Syn    & 98.6 & 1.9                         \\
\midrule
\multirow{3}{*}{\dict{}}               & Trans          & 95.7 & 1.1                         \\
                                        & Trans + Inv    & 64.2 & 9.2                         \\
                                        & Trans + Syn    & 93.7 & 2.2                         \\
\midrule
\multirow{3}{*}{\alignedmlm{}}            & Trans          & 98.4 & -0.5                         \\
                                        & Trans + Inv    & 95.2 & 6.7                         \\
                                        & Trans + Syn    & 98.2 & 3.6 \\
\bottomrule
\end{tabular}
\caption{Word embedding alignment scores for all pre-training methods considered. All four pretraining-methods use their default setup.}
\label{app:tab:correlation}
\end{table*}

From our results in Table~\ref{app:tab:correlation}, we observe that better word embedding alignment is strongly correlated with better transfer as measured by spearman correlation ($\rho_{s} = 0.727$, $p\textrm{ (2-tailed)}<.01$), and this is true for other tasks as well (NER -- $\rho_{s} = 0.781$, $p<.01$ and POS -- $\rho_{s} = 0.734$, $p<.01$).

\section{Transformations and their examples}
\label{app:transformations}

We present examples of transformations in Table~\ref{app:tab:trans_examples}, borrowed from~\citet{deshpande2021bert}.
We follow~\citet{deshpande2021bert} for all the transformations.
\paragraph{Transliteration}
We create a copy of the vocabulary for all the tokens in \orig{} other than the special tokens like \texttt{[CLS]}.
\paragraph{Inversion}
We invert the order of tokens in a sentence.
We use the period token to decide sentence boundaries both during pre-training and fine-tuning.
\paragraph{Syntax}
We modify the syntax of the English sentence to match that of French by using~\citet{wang2016galactic} which stochastically re-orders the dependents of nouns and verbs in the dependency parse.

\newcommand{\graytext}[1]{\textcolor{gray}{#1}}

\begin{table*}[t]
    \centering
    \begin{tabular}{@{}ll@{\hskip 3em}l@{}}
    \toprule
    \textbf{Transformation} & \textbf{Instance ($s$)}                                      & \textbf{Transformed instance $\left ( \textrm{\trans}(s) \right )$}                        \\ \midrule
    \textbf{\textit{Inversion}} (\inv) & Welcome to ACL                               & ACL to Welcome                             \\
    
    
    \textbf{\textit{Transliteration}} (\script) & I am Sam . I am                                        & $\clubsuit_\textrm{\graytext{(I)}}\; \heartsuit_\textrm{\graytext{(am)}}\; \diamondsuit_\textrm{\graytext{(Sam)}}\; \spadesuit_\textrm{\graytext{(.)}}\; \clubsuit_\textrm{\graytext{(I)}}\; \heartsuit_\textrm{\graytext{(am)}}\; $                                         \\ 
    
    \multirow{2}{*}{\textbf{\textit{Syntax}} (\syntax)} & Sara {\small \graytext{(S)}} ate {\small \graytext{(V)}} apples {\small \graytext{(O)}} & Sara {\small \graytext{(S)}} apples {\small \graytext{(O)}} ate {\small \graytext{(V)}} \\
    
     & Une table {\small \graytext{(N)}} ronde {\small \graytext{(A)}}  & Une ronde {\small \graytext{(A)}} table {\small \graytext{(N)}} \\
    
    \bottomrule
    \end{tabular}%
    \caption{
    Examples of our transformations applied to different sentences (without sub-word tokenization).
    \textit{Inversion} inverts the tokens, \textit{Permutation} samples a random reordering, and \textit{Transliteration} changes the script.
    We use symbols ($\clubsuit$) to denote words in the new script and mention the corresponding original word in brackets.
    \textit{Syntax} stochastically modifies the syntactic structure.
    In the first example for \textit{Syntax}, the sentence in Subject-Verb-Object (SVO) order gets transformed to SOV order, and in the second, the sentence in Noun-Adjective (NA) order gets transformed to the AN order.
    The examples are high probability re-orderings and other ones might be sampled too.
    }
    \label{app:tab:trans_examples}
\end{table*}

\section{Implementation Details}
\label{app:implementation}
\subsection{Data Generation}
\label{app:datageneration}
Our base English training corpus is the Wikitext-103 training corpus~\cite{merity2016pointer}, the same as the one used by~\citet{deshpande2021bert}. Let this pre-training corpus be denoted as $\mathcal{C}_{\text{\orig{}}}$. To generate the synthetic corpus, we perform the transformation on each sentence individually to generate a new corpus, which we will denote as $\mathcal{C}_{\text{\deriv{}}}$.
For the syntax transformation (\syntax{}), we follow~\citet{deshpande2021bert} and use galactic dependencies~\cite{wang2016galactic} to stochastically permute the nodes in the dependency parse from \orig{}'s syntax (here, English) to that of French.

The procedure for generating the finetuning corpora is the same. We perform the desired transformation on the input sentences, as well as any other corresponding labels. In XNLI, we also transform the second sentence. In NER and POS, we transform the tags of each word by moving them to the corresponding location where the actual word has been moved to. In NER, we do additional transformations on sequences of $\textsc{B}$ and $\textsc{O}$ tags to ensure \textsc{B} always comes before \textsc{O}, which is a requirement of the NER task.

Our vocabulary of size 40,000 is generated using the same approach as~\citet{deshpande2021bert}; we use the shared-byte-pair-encoding tokenizer described in~\cite{sennrich2015neural}. Bilingual dictionaries are generated using two sets of tokens created using transliteration, one for the original-language and one for the derived-language, with a 1-to-1 mapping between them. In particular, the mapping represents the translation between the words of the dictionary.

\subsection{\mlm{}}
\label{app:mlm}
Our approach for \mlm{} follows the same structure as~\citet{deshpande2021bert}. The only difference is that we perform syntax and inversion transformations on a word level, instead of on a token level. Their approach involved tokenizing the words, then performing the transformation on the tokens, whereas our approach involves first performing the transformation on the words of a sentence to generate $\mathcal{C}_{\text{\deriv{}}}$, before passing it into the \mlm{} model. For more information, see~\ref{app:datageneration}.

\subsection{\xlm{}}
\label{app:xlm}
Let us define the set of sentences in the original dataset as $s_{1:N}^\text{\orig{}}$, where $N$ is the number of sentences in $\mathcal{C}_{\text{\orig{}}}$. Similarly, the sentences in our synthetic language will be denoted as $s_{1:N}^\text{\deriv{}}$. Both are individually tokenized into $N$ sentences of word tokens, which we'll denote $x_{1:N}^\text{\orig{}}$ and $x_{1:N}^\text{\deriv{}}$. We keep concatenating these sentences of tokens together until including any more would exceed the max tokens per instance limit of $H=512$, where an instance represents a single vector of data being fed to the model (equivalently the coresponding data for a batch size of 1). Therefore, each instance is of the form $v_i^\mathcal{L} = [x_{a_{\mathcal{L}, i}:a_{\mathcal{L}, i+1}}^\mathcal{L}, \text{pad}_i^\mathcal{L}]$, where $\mathcal{L} \in \{\text{\orig{}}, \text{\deriv{}}\}$, $x_{a_{\mathcal{L}, i}:a_{\mathcal{L}, i+1}}^\mathcal{L}$ is the concatenation of the token sentences in $\mathcal{L}$ from $a_{\mathcal{L}, i}$ to $a_{\mathcal{L}, i+1}-1$ inclusive, and $\text{pad}_i^\mathcal{L}$ is a sequence of padding tokens to ensure $v_i^\mathcal{L}$ is of length 512. Simultaneously, as in all RoBERTa models, we generate corresponding position tokens for every input token in $v_i^\mathcal{L}$. We also assign a language ID token for every input token in $v_i^\mathcal{L}$, to indicate if the input token is from \orig{} or \deriv{}. The position and language ID tokens corresponding to padding word tokens are also changed to padding tokens.

Denote each combined instance of word, position, and language ID tokens as $w_i^\mathcal{L}$. Applying the above steps generates data of the form $\mathcal{D} = \{w_{1:M_1}^\text{\orig{}}, w_{1:M_2}^\text{\deriv{}}\}$, which we will denote as \mlm{} data since each instance is monolingual. It isn't necessarily the case that $M_1=M_2$, since the number of tokens in the original and synthetic text, even when parallel, can be different.

The next step is to generate \tlm{} data. Unlike the original implementation in \citet{lample2019cross}, the architecture we're using has a fixed maximum sequence length. Therefore, we decided to always divide an instance into (at most) 256 tokens of the original language followed by (at most) 256 instances of the synthetic language. The remaining space in both the first and second half is filled up with post-padding tokens. In every instance, we keep sampling until adding another parallel sentence of both original and synthetic data would cause the tokens in the original sentences or synthetic sentences to exceed 256 tokens. We then populate the instance with padding tokens as described above, before continuing with the next set of sentences. This results in $M_3$ \tlm{} instances of the form $w_{1:M_3}^\text{\tlm{}} = [x_{a_{\text{\orig{}}, i}:a_{\text{\orig{}}, i+1}}^\text{\orig{}}, \text{pad}_{1,i}^\tlm{}, x_{a_{\text{\deriv{}}, i}:a_{\text{\deriv{}}, i, i+1}}^\text{\deriv{}}, \text{pad}_{2,i}^\tlm{}]$. Finally, the \mlm{} and \tlm{} data are combined yielding our final dataset $\{w_{1:M_1}^\text{\orig{}}, w_{1:M_2}^\text{\deriv{}}, w_{1:M_3}^\text{\tlm{}}\}$.

In our \xlm{} model, for both \mlm{} and \tlm{} data, all sentences that exceed 512 or 256 tokens respectively are thrown out. Most sentences are significantly shorter than this, so few were actually thrown out. We decided this was a reasonable tradeoff to avoid the complications that arise from splitting a translated pair of sentences into multiple instances, while trying to ensure each resulting instance pair is still a translation of the other. Finally, note that for finetuning, we do not need to generate \tlm{} data.

We ran a number of ablation studies on \xlm{} by varying the amount of \tlm{} data relative to \mlm{}, and by varying the percentage of vocabulary overlap between \orig{} and \deriv{}. The results of these studies can be found in Appendix \ref{app:tlm_data} and \ref{app:vocabulary}. The default setup for \xlm{} uses $25\%$ \tlm{} data relative to \mlm{}, and no word overlap. This most closely resembles the ratio of \tlm{} and \mlm{} used in the original implementation of \xlm{} \cite{lample2019cross}.

\subsection{\dict{}}
We largely follow the \dict{}\textsc{-50} model introduced in \citet{dict-mlm} with two languages. With 50\% probability, the label of a masked token is altered according to the normal mBERT implementation. With the remaining 50\% probability, the masked token is changed to a cross-lingual synonym. Since we only work with 2 languages at a time, the latter case always changes the token to its synonym if it is in the bilingual dictionary; otherwise, we leave it as the original token. Finally, we did not include the language conditioning layer since its performance fluctuates compared to the vanilla \dict{} \cite{dict-mlm}.

Since \dict{} requires the use of language IDs, we built this pre-training method on top of our implementation of \xlm{}. The \dict{} objectives are applied onto the \mlm{} data, which we generate following Appendix \ref{app:xlm}. We do not generate \tlm{} data for \dict{}.

Our default setup uses 25\% of tokens in the bilingual dictionary. The original implementation of \dict{} uses the MUSE dataset, which contain 110 large-scale ground-truth bilingual dictionaries, some of which contain 100,000 token pairs \cite{conneau2017word, dict-mlm}. Since \orig{} only has a voacbulary size of 40,000 in our experiments, it's more than reasonable to have a dictionary capturing at least 25\% of the tokens. The effects of the percentage of tokens in the bilingual dictionary can be seen in Appendix \ref{app:dict_vs_aligned}.

\subsection{\alignedmlm{}}
Unlike \xlm{} and \dict{}, we don't want to have language ID embeddings in \alignedmlm{}. To see why, observe that in the RoBERTa model, language ID token embeddings would eventually be summed with the word token embeddings, before being passed through the model during finetuning and evaluation. However, the goal of \alignedmlm{} is to explicitly align embeddings between corresponding words in the bilingual dictionary, with no mention of the language ID embeddings. This means it is best not to include language ID tokens at all. Therefore, unlike \dict{}, \alignedmlm{} is directly constructed using the \mlm{} model explained in Appendix \ref{app:mlm}. The only difference is we additionally apply the objective function explained in Section \ref{sec:approach}, using an $\alpha$ of 10. $\alpha=10$ was chosen from a simple grid search on [1, 10, 100]. We noticed that the difference in performance was small and chose 10 because it performed marginally
better by 2 points on \script{} \comp{} \inv{}.

We ran ablation studies that varied the percentage of tokens contained in the bilingual dictionary. The default used was 25\%, and the full results can be found in Appendix \ref{app:dict_vs_aligned}.

\section{Hyperparameters}
\label{app:hyperparameters}
The model used in our experiments is a modified RoBERTa model \cite{liu2019roberta}. The parameters used during pretraining can be found in Table~\ref{app:tab:pretrain_hyperparameters}, while the parameters used during finetuning can be found in Table~\ref{app:tab:finetune_hyperparameters}.

\begin{table}[ht]
\resizebox{1\columnwidth}{!}{
\begin{tabular}{ll}
\toprule
\multicolumn{1}{l}{\bf{Parameter Name}}
& \multicolumn{1}{l}{\bf{Parameter Value}}\\
\midrule
Number of Attention Heads & 8\\
Number of Hidden Layers & 8\\
Hidden Dimensionality & 512\\
Training Steps & 500,000\\
Batch size & 128\\
Learning Rate & $10^{-4}$\\
Linear Warmup & 10,000\\
\lang1 Vocabulary Size & 40,000\\
\bottomrule
\end{tabular}
}
\caption{This is the list of pretraining parameters used in our experiments. Any unlisted parameters use the same values as in RoBERTa \cite{liu2019roberta}.}
\label{app:tab:pretrain_hyperparameters}
\end{table}

\begin{table}[ht]
\resizebox{1\columnwidth}{!}{
\begin{tabular}{clrrrrrrrrr}
\toprule
\multicolumn{1}{l}{\bf{Evaluation Benchmark}}
& \multicolumn{1}{l}{\bf{Parameter}} & \multicolumn{1}{l}{\bf{Value}} \\
\midrule
\multirow{4}{*}{XNLI}                    & Learning Rate          & $2 \times 10^{-5}$                          \\
                                        & Max Sequence Length    & 128                           \\
                                        & Training Epochs    & 5                          \\
                                        & Batch Size    & 32                          \\
\midrule
\multirow{4}{*}{NER}                    & Learning Rate          & $2 \times 10^{-5}$                          \\
                                        & Max Sequence Length    & 128                           \\
                                        & Training Epochs    & 10                          \\
                                        & Batch Size    & 32                          \\
\midrule
\multirow{4}{*}{POS}                    & Learning Rate          & $2 \times 10^{-5}$                          \\
                                        & Max Sequence Length    & 128                           \\
                                        & Training Epochs    & 10                          \\
                                        & Batch Size    & 32                          \\
\bottomrule
\end{tabular}
}
\caption{Finetuning parameters used in our experiments.}
\label{app:tab:finetune_hyperparameters}
\end{table}

\section{Training Time}
\label{app:training_time}

All four methods have similar training times. Using a v3-8 Google TPU on Google’s
Cloud platform, pretraining for 500,000 steps on \script{} \comp{} \inv{} took 40.7, 40.6, 44.9, and 39.7 hours using MLM, TLM, Align-MLM, and DICT-MLM respectively. These results are consistent across other transformations.

\end{document}